\documentclass[runningheads]{llncs}
\usepackage{graphicx}
\usepackage{amsmath,amssymb} 
\usepackage{color}
\usepackage{epsfig}
\usepackage{xspace}

\def \ie {\emph{i.e.}\xspace}
\def \eg {\emph{e.g.}\xspace}

\def \etal {\emph{et. al.}\xspace}
\def \vs {\emph{v.s.}\xspace}
\def \iset {\mathcal I}
\def \sset {\mathcal S}
\def \rset {\mathcal R}
\def \xset {\mathcal X}

\newcommand{\cm}{\checkmark}

\newcommand{\para}[1]{\noindent \textbf{#1}}

\DeclareMathAlphabet{\mathpzc}{T1}{pzc}{m}{n}

\newlength{\halfwidth}
\newlength{\fullwidth}
\newlength{\tikzimgheight}
\newlength{\tikzimgwidth}

\usepackage{ifthen}

\usepackage[width=122mm,left=12mm,paperwidth=146mm,height=193mm,top=12mm,paperheight=217mm]{geometry}
\usepackage{multirow}
\usepackage{subfigure}
\usepackage{caption}
\usepackage{array}
\usepackage{booktabs}

\usepackage[numbers,sort&compress,square]{natbib} 

\makeatletter

\renewcommand\@biblabel[1]{#1.} 

\makeatother

\begin{document}
\pagestyle{headings}
\mainmatter

\title{Weakly Supervised Object Localization \\ Using Size Estimates} 

\titlerunning{Weakly Supervised Object Localization Using Size Estimates}

\authorrunning{Miaojing Shi and Vittorio Ferrari}

\author{Miaojing Shi and Vittorio Ferrari}


\institute{	University of Edinburgh\\
	\email{ \{miaojing.shi,vittorio.ferrari\}@ed.ac.uk}
}

\maketitle

\begin{abstract}

We present a technique for weakly supervised object localization (WSOL),
building on the observation that WSOL algorithms usually work better on images with bigger objects.
Instead of training the object detector on the entire training set at the same time, we propose a curriculum learning strategy to feed training images into the WSOL learning loop in an order from images containing bigger objects down to smaller ones. To automatically determine the order, we train a regressor to estimate the size of the object given the whole image as input.
Furthermore, we use these size estimates to further improve the re-localization step of WSOL by assigning weights to object proposals according to how close their size matches the estimated object size.
We demonstrate the effectiveness of using size order and size weighting on the challenging PASCAL VOC 2007 dataset, where we achieve a significant improvement over existing state-of-the-art WSOL techniques.


\end{abstract}



\vspace{-0.6cm}
\section{Introduction}

Object class detection has been intensively studied during recent years~\cite{dalal05cvpr,everingham10ijcv,felzenszwalb10pami,girshick15iccv,girshick14cvpr,MalisiewiczICCV11,uijlings13ijcv,viola:nips05,wang13iccv}.
The goal is to place a bounding box around every instance of a given object class.
Given an input image, typically modern object detectors first extract object proposals ~\cite{alexe10cvpr,uijlings13ijcv,dollar14eccv} and then score them with a classifier to determine their probabilities of containing an instance of certain class~\cite{cinbis14cvpr,cinbis15pami}.
Manually annotated bounding boxes are typically required for training the classifier.

Annotating bounding boxes is usually tedious and time consuming. In order to reduce the annotation cost,
a commonly used strategy is to learn the detector in a weakly supervised manner: we are given a set of images known to contain instances of a certain object class, but we do not know the object locations in these images. This weakly supervised object localization (WSOL) bypasses the need for bounding box annotation and therefore substantially reduces annotation time.
WSOL is typically conducted in two iterative steps~\cite{bilen14bmvc,bilen15cvpr,cinbis15pami,deselaers10eccv,russakovsky12eccv,siva11iccv,song14icml,song14nips}:
1) re-localizing object instances in the images using the current object detector, and
2) re-training the object detector given the current selection of instances.

\begin{figure*}[t]
\center
\includegraphics[width=1\columnwidth]{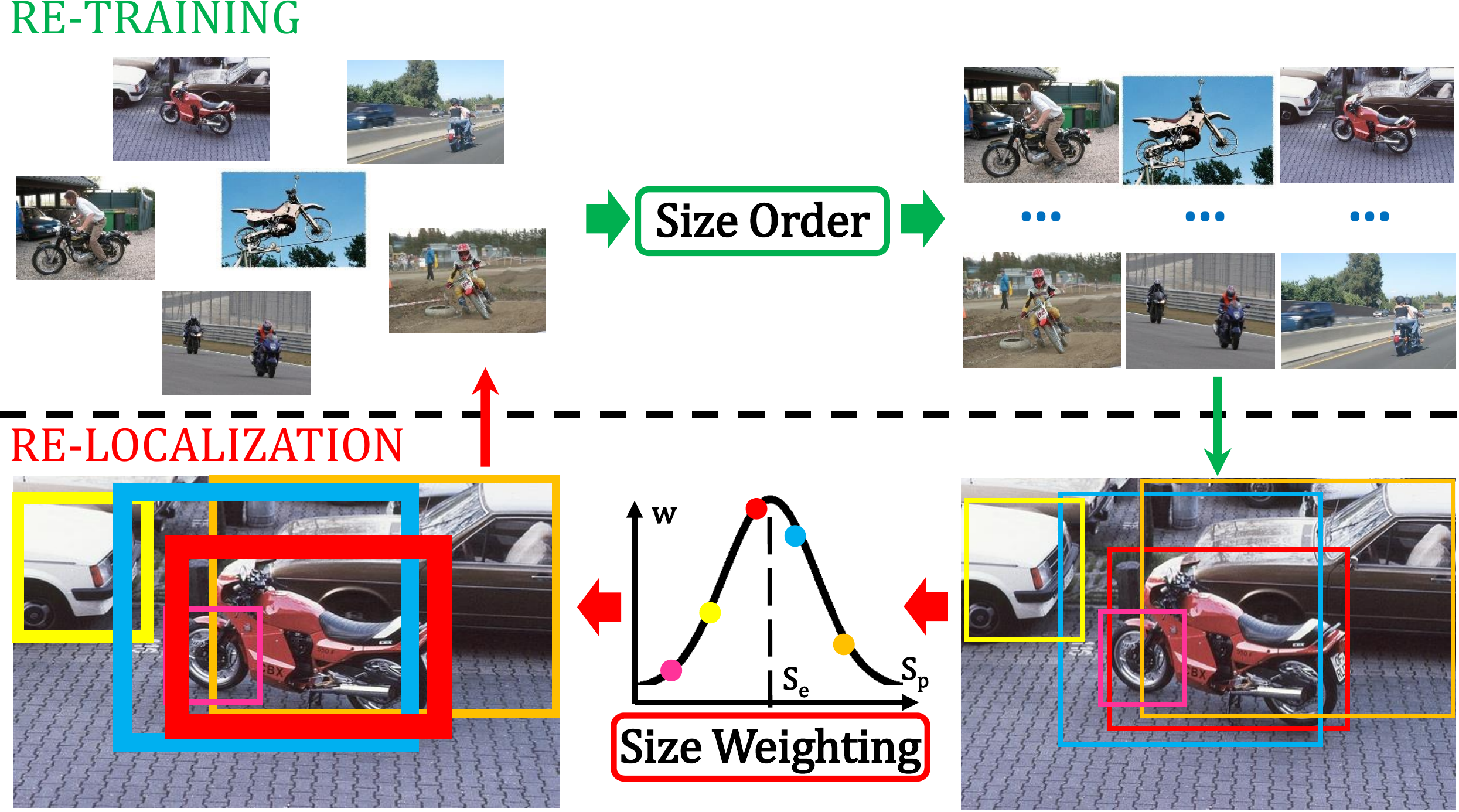}
\caption{\small Overview of our method. We use size estimates to determine the order in which images are fed to a WSOL loop,
so that the object detector is re-trained progressively from images with bigger objects down to smaller ones.
We also improve the re-localization step, by weighting object proposals according to how close their size ($s_p$) matches the estimated object size ($s_e$).
}
\label{Fig:Overview}
\end{figure*}

WSOL algorithms typically apply both the re-training and re-localization steps on the entire training set at the same time.
However, WSOL works better on images with bigger objects. For instance,~\cite{deselaers10eccv} observed that the performance of several WSOL 
algorithms consistently decays from easy dataset with many big objects (Caltech4~\cite{fergus2003cvpr}) to hard dataset with many 
small objects (PASCAL VOC 07~\cite{everingham10ijcv}). 
In this paper, we propose to feed images into the WSOL learning loop in an order from images containing bigger objects down to smaller ones (Fig.~\ref{Fig:Overview}, top half). This forms a curriculum learning~\cite{bengio09icml} strategy where the learner progressively sees more and more training samples, starting from easy ones (big objects) and gradually adding harder ones (smaller objects).
To understand why this might work better than standard orderless WSOL, let's compare the two. 
The standard approach re-trains the model from {\em all} images at each iteration. These include many incorrect localizations which corrupt the model re-training,
and result in bad localizations in the next re-localization step, particularly for small objects (Fig.~\ref{Fig:currculum_learning}).
In our approach instead, WSOL learns a decent model from images of big objects in the first few iterations. This initial model then better localizes objects in images of mid-size objects, which in turn leads to an even better model in the next re-training step, as it has now more data, and so on. By the time the process reaches images of small objects, it already has a good detector, which improve the chances of localizing them correctly (Fig.~\ref{Fig:currculum_learning}).

Our easy-to-hard strategy needs to determine the sequence of images automatically. For this we train a regressor to estimate the size of the object given the whole image as input. In addition to establishing a curriculum, we use these size estimates to improve the re-localization step. We weight object proposals according to how close their size matches the estimated object size (Fig.~\ref{Fig:Overview}, bottom half). These weights are higher for proposals of size similar to the estimate, and decrease as their size difference increases. This weighting scheme reduces the uncertainty in the proposal distribution, making the re-localization step more likely to pick a proposals correctly covering the object. Fig.~\ref{Fig:score_function} shows an example of how size weighting changes the proposal score distribution induced by the current object detector, leading to more accurate localization.

In extensive experiments on the popular PASCAL VOC 2007 dataset, we show that:
1) using our curriculum learning strategy based on object size gives a 7\% improvement in CorLoc compared to the orderless WSOL;
2) by further adding size weighting into the re-localization step, we get another 10\% CorLoc improvement;
3) finally, we employ a deep Neural Network to re-train the model and achieve our best performance, significantly outperforming the state-of-the-art in WSOL~\cite{cinbis15pami,wang15tip,bilen15cvpr}.

Compared to standard WSOL, our scheme needs additional data to train the size regressor. This consists of a single scalar value indicating the size of the object, for each image in an external dataset. We do not need bounding-box annotation. Moreover, in Sec.~\ref{Sec:further_analysis} we show that we can use a size regressor generic across classes, by training it on different classes than those used during WSOL.

\section{Related Work}


\para{Weakly-supervised object localization (WSOL).}
In WSOL the training images are known to contain instances of a certain object class but their locations are unknown. The task is both to localize the objects in the training images and to learn an detector for the class.
WSOL is often conceptualised as Multiple Instance Learning
(MIL)~\cite{bilen14bmvc,cinbis14cvpr,deselaers10eccv,dietterich97ai,shi12bmvc,siva11iccv,song14icml,song14nips}.
Images are treated as bags of object proposals~\cite{alexe10cvpr,uijlings13ijcv,dollar14eccv} (instances). A negative image contains only negative instances. A positive image contains at least one positive instance, mixed in with a majority of
negative ones. The goal is to find the true positives instances from which to learn a classifier for the object class.

Due to the use of strong CNN features~\cite{girshick14cvpr,krizhevsky12nips}, recent works on WSOL~\cite{bilen14bmvc,bilen15cvpr,cinbis14cvpr,song14icml,song14nips,wang15tip} have shown
remarkable progress.
Moreover, researchers also tried to incorporate various advanced cues into the WSOL process, \eg objectness~\cite{cinbis15pami,deselaers10eccv,siva11iccv,tang14cvpr,alexe12pami},
co-occurrence between multiple classes in the same training images~\cite{shi12bmvc},
and even appearance models from related classes learned from bounding-box annotations~\cite{guillaumin12cvpr,rochan2015cvpr,hoffman14nips}. 
In this work, we propose to estimate the size of the object in an image and inject it as a new cue into WSOL.
We use it both to determine the sequence of training images in a curriculum learning scheme, and to weight the score function used during the re-localization step.

\para{Curriculum learning (CL).}
The curriculum learning paradigm was proposed by Bengio \etal~\cite{bengio09icml}, in which the model was learnt gradually from easy to hard samples so as to increase the entropy of training. A strong assumption in~\cite{bengio09icml} is that the curriculum is provided by a human teacher. In this sense, determining what constitute an easy sample is subjective and needs to be manually provided. To alleviate this issue, Kumar and Koller~\cite{kumar2010nips} formulated CL as a regularization term into the learning objective and proposed a self-paced learning scheme.

The concept of learning in an easy-to-hard order was visited also in computer vision~\cite{lee11cvpr,pentina2015cvpr,sharmanska2013cvpr,lapin2014nn,ionescu16cvpr}.
These works focus on a key question: what makes an image easy or hard?
The works differ by how they re-interpret ``easiness" in different scenarios.
Lee and Grauman~\cite{lee11cvpr} consider the task of discovering object classes in an unordered image collection. They relate easiness to ``objectness" and ``context-awareness".
Their context-awareness model is initialized with regions of ``stuff" categories, and is then used to support discovering ``things" categories in unlabelled images.
The model is updated by identifying the easy object categories first and progressively expands to harder categories.  
Sharmanska \etal~\cite{sharmanska2013cvpr} use some privileged information to distinguish between easy and hard examples in an image classification task. The privileged information are additional cues available at training time, but not at test time. They employ several additional cues, such as object bounding boxes, image tags and rationales to define their concept of easiness~\cite{lapin2014nn}.
Pentina \etal~\cite{pentina2015cvpr} consider learning the visual attributes of objects.
They let a human decide whether an object is easy or hard to recognize. The human annotator provides a difficulty score for each image, ranging from easy to hard.
In this paper, we use CL in a WSOL setting and propose object size as an ``easiness" measure.
The most related work to ours is the very recent~\cite{ionescu16cvpr}, which learns to predict human response times as a measure of difficulty, and shows an example application to WSOL.




\section{Method}

In this section we first describe a basic MIL framework, which we use as our baseline (Sec.~\ref{Sec:shallow_MIL}); then we show how to use object size estimates to improve the basic framework by introducing a sequence during re-training (Sec.~\ref{Sec:size_order}) and a weighting during re-localization (Sec.~\ref{Sec:size_weighting}). Finally, we explain how to obtain size estimates automatically in Sec.~\ref{Sec:size_regressor}.

\begin{figure*}[t]
\center
\includegraphics[width=1\columnwidth]{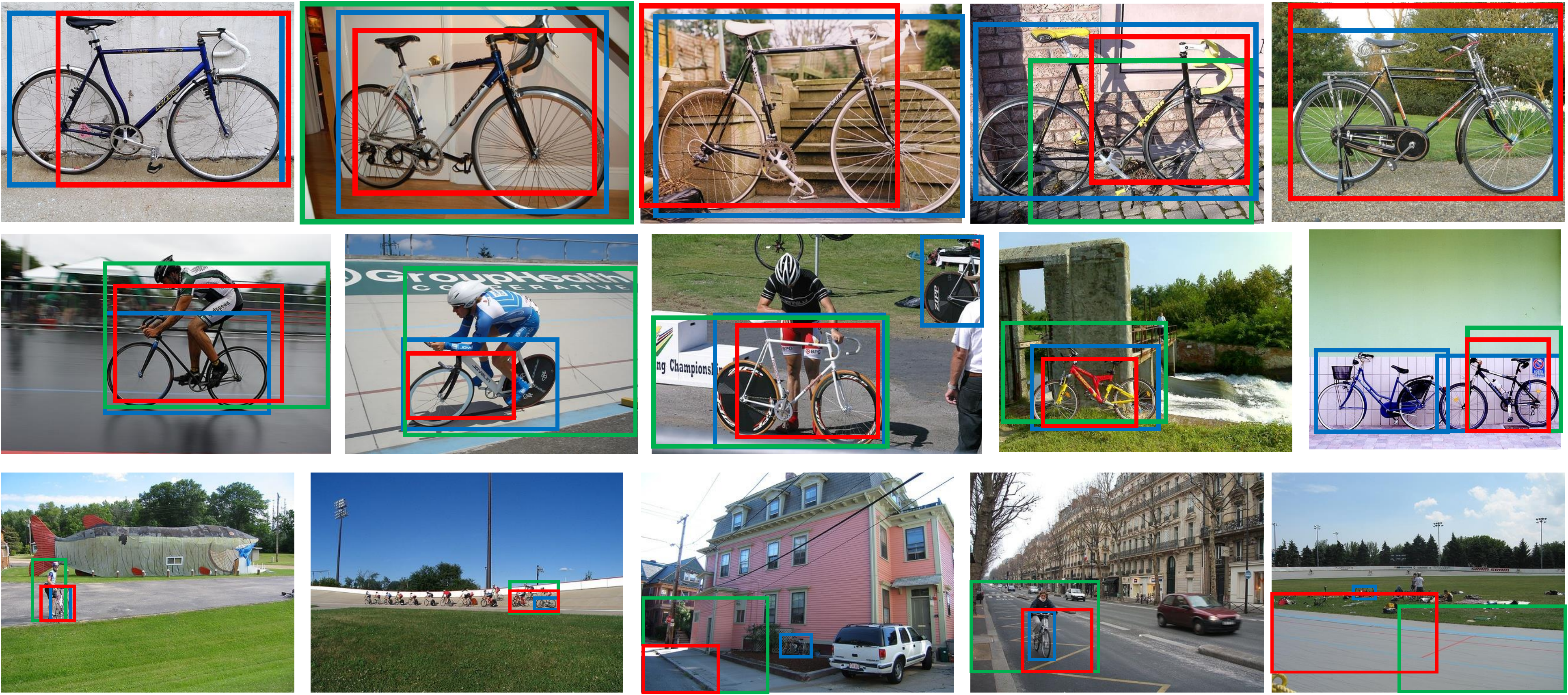}
\caption{\small Illustration of the estimated size order for class \emph{bicycle}, for three batches (one per row).
We show the ground-truth object bounding-boxes (blue), objects localized by our WSOL scheme using size order (red),
and objects localized by the basic MIL framework (green). In the first, third and last examples of the first row the green and red boxes are identical.}
\label{Fig:currculum_learning}
\end{figure*}

\subsection{Basic Multiple Instance Learning framework}\label{Sec:shallow_MIL}

We represent each image in the input set $\iset$ as a bag of proposals extracted using the state of the art
object proposal method~\cite{dollar14eccv}. It returns about 2000 proposals per image, likely to cover all objects.
Following~\cite{girshick14cvpr,bilen14bmvc,song14icml,song14nips,wang15tip},
we describe the proposals by the output of the second-last layer of the CNN model proposed by Krizhevsky \etal~\cite{krizhevsky12nips}.
The CNN model is pre-trained for whole-image classification on ILSVRC~\cite{russakovsky15ijcv}, using the Caffe implementation~\cite{jia13caffe}.
This produces a 4096-dimensional feature vector for each proposal.
Based on this feature representation, we iteratively build an SVM appearance model $A$ (object detector) in two alternating steps:
(1) re-localization: in each positive image, we select the highest scoring proposal by the SVM. This produces the set $\sset$ which contains the current selection of one instance from each positive image.
(2) re-training: we train the SVM using $\sset$ as positive training samples, and all proposals from the negative images as negative samples.

As commonly done in~\cite{cinbis15pami,cinbis14cvpr,russakovsky12eccv,pandey11iccv,nguyen09iccv,kim2009nips}
we initialize the process by training the appearance model using complete images as training samples. Each image in $\iset$ provides a training sample. 
Intuitively, this is a good initialization when the object covers most of the image, which is true only for some images.
%

\begin{table}[t]
\renewcommand{\arraystretch}{0.8} 
\centering
 \small{ \label{Alg:algorithm} {\small
     \begin{tabular}{m{1\columnwidth}}
     \toprule
     \textbf{Alg. 1} Multiple instance learning with size order and size weighting \\
     \midrule
      \textbf{Initialization:} \\
      1) split the input set $\iset$ into $K$ batches according to the estimated object size order\\
      2) initialize the positive and negative examples as the entire images in first batch $\mathcal{I}_1$ \\
      3) train an appearance model $A_1$ on the initial training set\\
      \textbf{for} batch $k = 1:K$ \textbf{do}\\
      \setlength{\parindent}{2em}  \textbf{for} iteration $m = 1:M$ \textbf{do} \\
       \setlength{\parindent}{4em}  i)~~~\textbf{re-localize} the object instances in images $\cup_{i=1}^{k} \mathcal{I}_i$ using current appearance \\
       \setlength{\parindent}{4em} ~~~~~model $A_{k}^m$ and size weighting of object proposals;\\
       \setlength{\parindent}{4em}
       ii)~~add new negative proposals by hard negative mining;\\
       \setlength{\parindent}{4em}
       iii) \textbf{re-train} the appearance model $A_{k}^m$ given current selection of instances in \\
       \setlength{\parindent}{4em} ~~~~~images $\cup_{i=1}^{k} \mathcal{I}_i$;\\
         \setlength{\parindent}{2em}    \textbf{end for} \\
         \textbf{end for} \\
       Return final detector and selected object instances in $\iset$.\\
           \bottomrule
    \end{tabular}
} }
\end{table}

\subsection{Size order}
\label{Sec:size_order}
Assume we have a way to automatically estimate the size of the object in all input images $\mathcal{I}$ (Sec.~\ref{Sec:size_regressor}). Based on their object size order, we re-organize MIL on a curriculum, as detailed in Alg. 1.

We split the images into $K$ batches according to their estimated object size (Fig.~\ref{Fig:currculum_learning}).
We start by running MIL on the first batch $\mathcal{I}_1$, containing the largest objects. The whole-image initialization works well on them, leading to a reasonable first appearance model $A_1$ (though trained from fewer images).
We continue running MIL on the first batch $\mathcal{I}_1$ for $M$ iterations to get a solid $A_1$. %
The process then moves on to the second batch $ \mathcal{I}_2$, which contains mid-size objects, {\em adding} all its images into the current working set $\mathcal{I}_1 \cup \mathcal{I}_2$, and run the MIL iterations again.
Instead of starting from scratch, we use $A_1$ from the first batch MIL iterations. This model is likely to do a better job at localizing objects in batch $\mathcal{I}_2$ than the whole-image initialization of basic MIL (Fig.~\ref{Fig:currculum_learning}, second row). Hence, the model trains from better samples in the re-training step. Moreover, the model $A_2$ output by MIL on $\mathcal{I}_1 \cup \mathcal{I}_2$ will be better than $A_1$, as it is trained from more samples. Finally, during MIL on $\mathcal{I}_1 \cup \mathcal{I}_2$, the localization of objects in $I_1$ will also improve (Fig.~\ref{Fig:currculum_learning}, first row).

The process iteratively moves on to the next batch $k+1$, every time starting from appearance model $A_k$ and running MIL's re-training / re-localization iterations on the image set $\cup_{i=1}^{k+1} \mathcal{I}_i$.
As the image set continuously grows, the process does not jump from batch to batch. This helps stabilizing the learning process and properly training the appearance model from more and more training samples.
By the time the process reaches batches with small objects, the appearance model will already be very good and will do a much better job than the whole-image initialization of basic MIL on them (Fig.~\ref{Fig:currculum_learning}, third row).
Fig.~\ref{Fig:currculum_learning} shows some examples of applying our curriculum learning strategy compared to basic MIL. In all our work, we set $K = 3$ and $M = 3$.

%

\subsection{Size weighting}
\label{Sec:size_weighting}

In addition to establishing a curriculum, we use the size estimates to refine the re-localization step of MIL.
A naive way would be to filter out all proposals with size different from the estimate.
However, this is likely to fail as neither the size estimator nor the proposals are perfectly accurate, and therefore even a good proposal covering the object tightly will not exactly match the estimated size.

Instead, we use the size estimate as indicative of the {\em range} of the real object size. Assuming the error distribution of the estimated size w.r.t the real size is normal, according to the three-sigma rule of thumb~\cite{wheeler1992understanding}, the real object size is very likely to lie in this range $[s_e - 3\sigma, s_e + 3\sigma]$ (with 99.7\% probability),
where $s_{e}$ is the estimated size and $\sigma$ is the standard deviation of the error.
We explain in Sec.~\ref{Sec:size_regressor} how we obtain $\sigma$.

\begin{figure}[t]
 \centering
 \subfigure{\includegraphics[width=0.40\columnwidth]{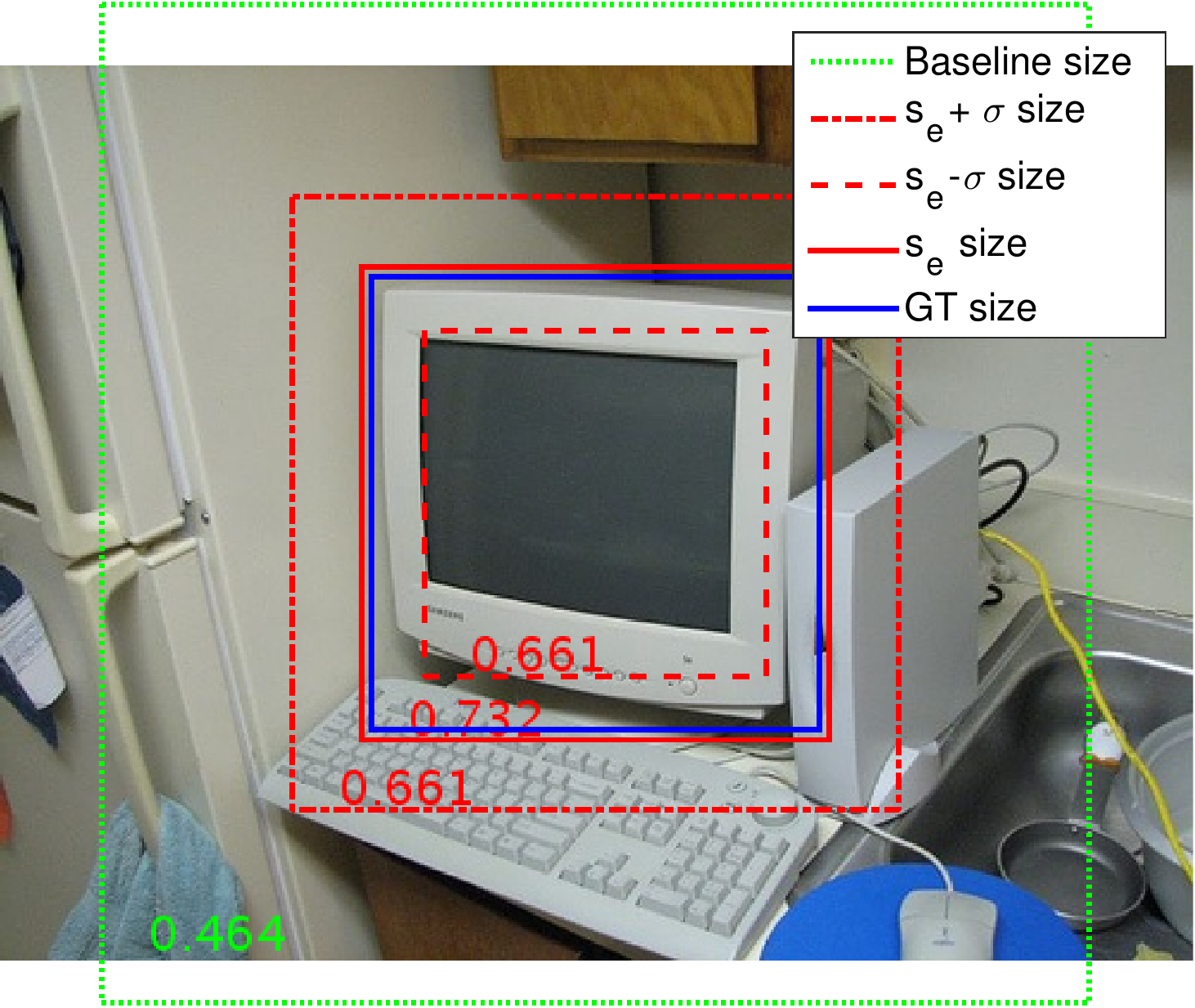}}
 \subfigure{\includegraphics[width=0.45\columnwidth]{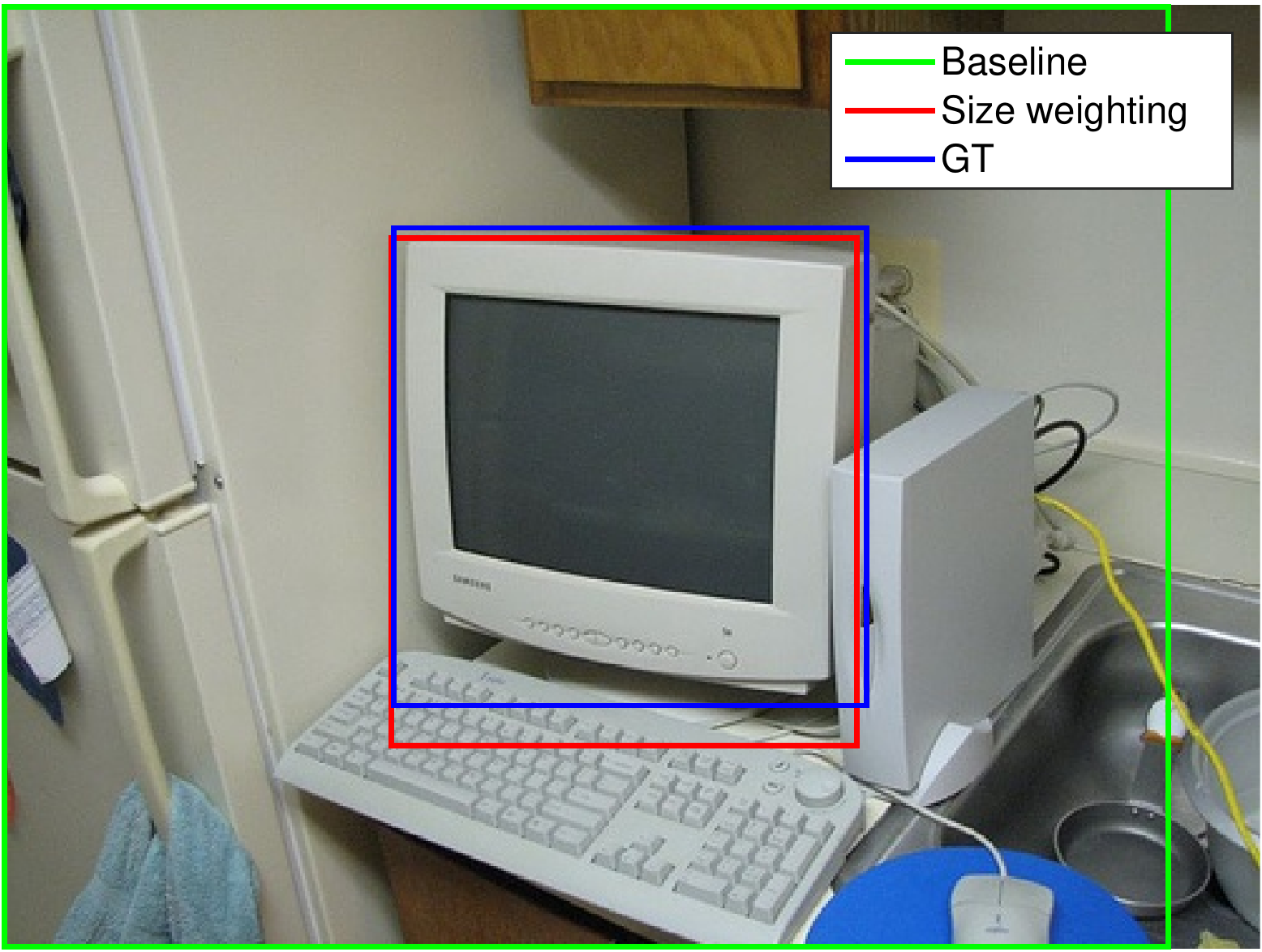}}
 \caption{\small{Illustration of size weighting.
 Left: behaviour of the size weighting function $W$. Example sizes are shown by boxes of the appropriate area centered at the ground truth (GT) object; $s_e$ denotes the estimated object size. The size weight $W$ of each box is written in its bottom left corner.
 Right: detection result using size weighting (red) compared to basic MIL framework (green).}}
\label{Fig:score_function}
\end{figure}

We assign a continuous weight to each proposal $p$ so that it gives a relatively high weight for the size $s_p$ of the proposal falling inside the $3\sigma$ interval of the estimated object size $s_e$,
and a very low weight for $s_p$ outside the interval:
\begin{equation}\label{Eqn:size_score}
    W(p; s_e, \sigma, \delta ) = \min\left(\frac{1}{{1 + {e^{\delta \cdot (s_e -3\sigma - s_p)}}}}  \; , \; \frac{1}{{1 + {e^{\delta \cdot (s_p - s_e - 3\sigma)}}}}\right).
\end{equation}
This function decreases with the difference between $s_p$ and $s_e$ (Fig.~\ref{Fig:score_function});
$\delta$ is a scalar parameter that controls how rapidly the function decreases, particularly outside the three sigma range $[s_l, s_r]$.
The model is not sensitive to the exact choice of $\delta$ (we set $\delta=3$ in all experiments). Weights for proposals falling out of the interval $[s_l, s_r]$ quickly go to zero.
Thereby this weight $W$ represents the likelihood of proposal $p$ covering the object, according to the size estimate $s_e$.

We now combine the size weighting $W$ of a proposal with the score given by the SVM appearance model $A$.
First we transform the output of the SVM into a probability using platt-scaling~\cite{platt1999}.
Assuming that the two score functions are independent, we combine them by multiplication, yielding the final score of a proposal $p$: $ A(p)  \cdot  W(p; s_e, \sigma, \delta ) $.
This score is used in the re-localization step of MIL (Sec.~\ref{Sec:shallow_MIL}), making it more likely to pick a proposal correctly covering the object. Fig.~\ref{Fig:result} gives some example results of using this size weighting model.



\subsection{Size estimator}
\label{Sec:size_regressor}
In subsections~\ref{Sec:size_order} and~\ref{Sec:size_weighting}, we assumed the availability of an automatic estimator of the size of objects in images. In this subsection we explain how we do it.

We use Kernel Ridge Regressor (KRR)~\cite{shawe2004kernel} to estimate the size of the object given the whole image as input. We
train it beforehand on an external set $\rset$, disjoint from the set $\mathcal{I}$ on which MIL operates (Sec.~\ref{Sec:shallow_MIL}).
We train a separate size regressor for each object class. For each class, the training set $\rset$ contains images annotated with the size $s_t$ of the largest object of that class in it. The training set can be small, as we demonstrate in Sec.~\ref{Sec:Details_analysis}.
The input image is represented by a 4096-dimensional CNN feature vector covering the whole image, output of the second-last layer of the AlexNet CNN architecture~\cite{krizhevsky12nips}.
The object size is represented by its area normalized by the image area.
As area differences grow rapidly, learning to directly regress to area puts more weight on estimation errors on large objects rather than on smaller objects.
To alleviate this bias, we apply a $r$-th root operation on the regression target values $s_t \leftarrow \sqrt[r]{s_t}$. Empirically, we choose $r=3$, but the regression performance over different $r$ is very close.

We train the KRR by minimizing the squared error on the training set $\rset$ and obtain the regressor along with the standard deviation $\sigma$ of its error by cross-validation on $\rset$.
We then use this size regressor to automatically estimate the object size on images in the WSOL input set $\iset$.

\section{Experiments}

\subsection{Dataset and settings}

\para{Size estimator training.}
We train the size estimator on the trainval set $\rset$ of PASCAL VOC 2012~\cite{everingham10ijcv} (PASCAL 12 for short). This has 20 classes, a total of 11540 images, and 834 images per class on average.

\para{WSOL.}
We perform WSOL on the trainval set $\iset$ of PASCAL 07~\cite{everingham10ijcv}, which has different images of the same 20 classes in $\rset$ (5011 images in total). While several WSOL works remove images containing only truncated and difficult objects~\cite{cinbis14cvpr,cinbis15pami,deselaers10eccv,russakovsky12eccv}, we use the complete set $\iset$.

We apply the size estimator on $\iset$ and evaluate its performance on it in Sec.~\ref{Sec:Preliminary}. Then, we use the estimated object sizes to improve the basic MIL approach of Sec.~\ref{Sec:shallow_MIL}, as described in Sec.~\ref{Sec:size_order}~and~\ref{Sec:size_weighting}.
Finally, we apply the detectors learned on $\iset$ to the test set $\xset$ of PASCAL 07, which contains 4952 images in total.
We evaluate our method and compare to standard orderless MIL in Sec.~\ref{Sec:MIL}.

\para{CNN.}
We use AlexNet as CNN architecture~\cite{krizhevsky12nips} to extract features for both size estimation and MIL (Sec.~\ref{Sec:shallow_MIL} and~\ref{Sec:size_regressor}). As customary~\cite{cinbis15pami,wang15tip,bilen15cvpr,song14nips}, we pre-train it for whole-image classification on ILSVRC~\cite{russakovsky15ijcv}, but we do {\em not} do any fine-tuning on bounding-boxes.

\subsection{Size estimation}
\label{Sec:Preliminary}

\para{Evaluation protocol.}
We train the regressor on set $\rset$. We adopt 7-fold cross-validation to obtain the best regressor and the corresponding $\sigma$.
In order to test the generalization ability of the regressor, we gradually reduce the number of training images from an average of 834 per class to 100, 50, 40, 30 per class.

The regression performance on $\iset$ is measured via the mean square error (MSE) between the estimated size and the ground-truth size (both in $r^{th}$ root, see Sec.~\ref{Sec:size_regressor}), and the Kendall's $\tau$ rank correlation coefficient~\cite{Kendall83} between the estimated size order and the ground-truth size order.


\medskip

\para{Results.}
Table~\ref{Tab:regression} presents the results.
We tried different $r^{\text{th}}$ root of the size value during training.
While $r = 3$ gives highest performance, it is not sensitive to exact choice of $r$, as long as $r>1$.
The table also shows the effect of reducing the number of training images $N$ to 100, 50, 40, and 30 per class. Although performance decreases when training with fewer samples, even using as few as 30 samples per class still delivers good results.

We set $r=3$ and use all training samples in $\rset$ by default in the following experiments.
We will also present an in-depth analysis of the impact of varying $N$ on WSOL in Sec~\ref{Sec:Details_analysis}.

\begin{table}[t]
\caption{\small Size estimation result on set $\iset$ with different $r$ and number $N$ of training images per class. $r$ refers to the $r^{\text{th}}$ root on size value applied;
`ALL' indicates using the complete $\rset$ set, which has 834 images per class on average. \label{Tab:regression}}
\begin{center}
   \begin{tabular}{|c|c||c|c|c|}
        \hline
   \multirow{2}{*}{$r^{\text{th}}$ root}&{Kendall's $\tau$}& \multirow{2}{*}{$N$}& {Kendall's $\tau$} & MSE\\
      \cline{2-2}
      \cline{4-5}
      & $N$  & & \multicolumn{2}{|c|}{$r =3$}\\
     \hline
    1  & 0.604 & \bf{ALL} & \bf{0.614} & \bf{0.013} \\
        \hline
    2 & 0.612& 100& 0.561& 0.016\\
        \hline
     \textbf{3} &\textbf{0.614} & 50& 0.542 & 0.018  \\
   \hline
   4  & 0.612 & 40 & 0.530 & 0.019\\
    \hline
    $5$ & $0.610$& 30 & 0.527 & 0.020\\
   \hline
    \end{tabular}
\end{center}
\end{table}

\subsection{Weakly supervised object localization (WSOL)}\label{Sec:MIL}

\para{Evaluation protocol.}
In standard MIL, given the training set $\iset$ with image-level labels, our goal is to localize the object
instances in this set and to train good object detectors for the test set $\xset$.
We quantify localization performance in the training set with the Correct Localization (CorLoc)
measure~\cite{bilen15cvpr,cinbis14cvpr,cinbis15pami,deselaers10eccv,shi15pami,wang15tip}.
CorLoc is the percentage of images in which the bounding-box returned by the algorithm correctly localizes an object of the target class (intersection-over-union $\geq 0.5$~\cite{everingham10ijcv}). We quantify object detection performance on the test set $\xset$ using mean average precision (mAP), as standard in PASCAL VOC.

As in most previous WSOL methods~\cite{bilen14bmvc,bilen15cvpr,cinbis14cvpr,cinbis15pami,deselaers10eccv,russakovsky12eccv,siva11iccv,song14icml,song14nips,wang15tip}, our scheme returns exactly one bounding-box per class per training image.
This enables clean comparisons to previous work in terms of CorLoc on the training set $\iset$.
Note that at test time the object detector is capable of localizing multiple objects of the same class in the same image
(and this is captured in the mAP measure).

\medskip

\para{Baseline.}
We use EdgeBoxes~\cite{dollar14eccv} as object proposals and follow the basic MIL framework of Sec.~\ref{Sec:shallow_MIL}.
For the baseline, we randomly split the training set $\iset$ into three batches ($K = 3$), then train an SVM appearance model sequentially batch by batch. We apply three MIL iterations ($M=3$) within each batch, and use hard negative mining for the SVM~\cite{cinbis14cvpr}.

Like in~\cite{cinbis15pami,deselaers10eccv,guillaumin12cvpr,prest12cvpr,shapovalova12eccv,siva11iccv,shi12bmvc,tang14cvpr,wang15tip}, we combine the SVN score with a general measure of ``objectness''~\cite{alexe10cvpr}, which measures how likely it is that a proposal tightly encloses an object of any class (\eg bird, car, sheep), as opposed to background (\eg sky, water, grass). For this we use the objectness measure produced by the proposal generator~\cite{dollar14eccv}.
Using this additional cue makes the basic MIL start from a higher baseline.

\begin{figure}[t]
\center
\includegraphics[width=1\columnwidth]{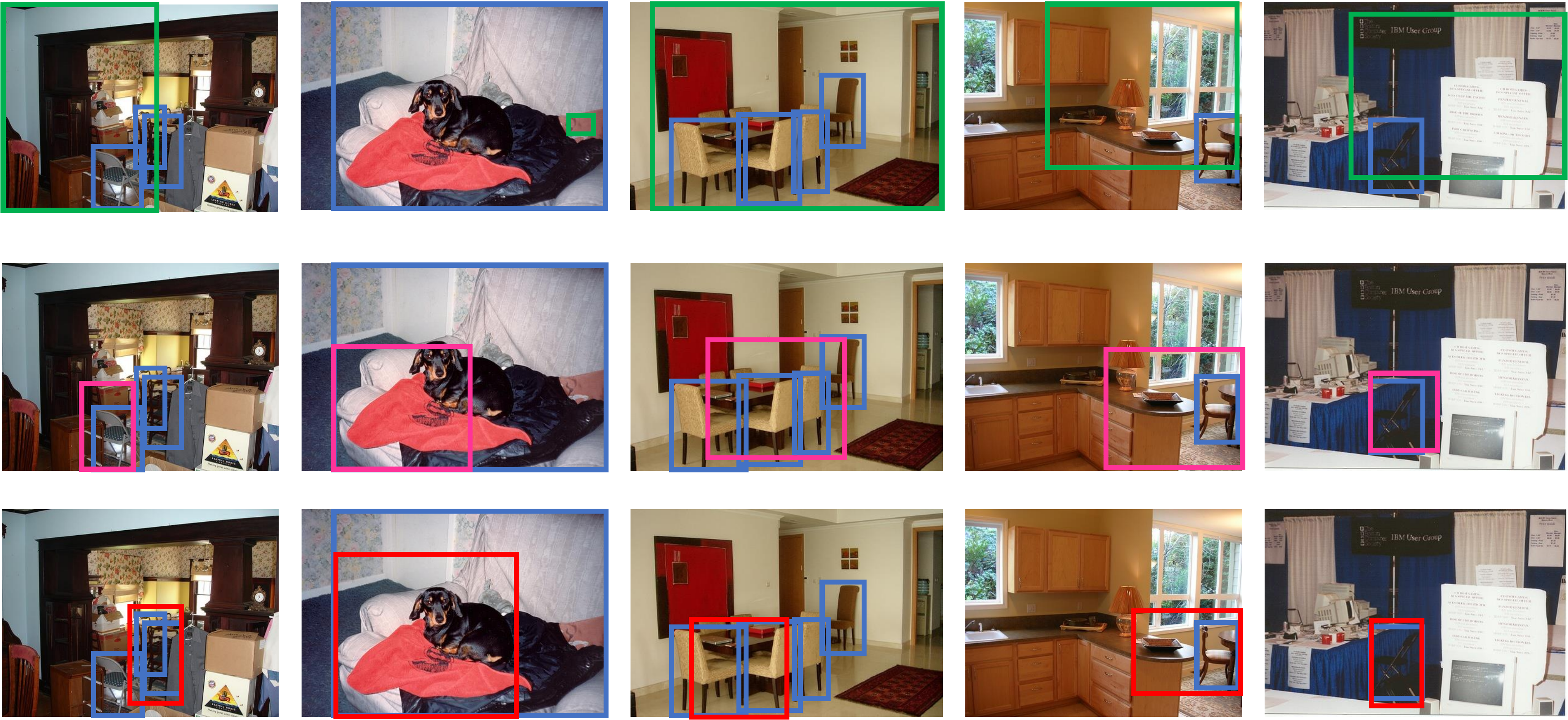}
\caption{\small Example localizations by different WSOL schemes on class \emph{chair}.
First row: localizations by the MIL baseline (green, see Sec.~\ref{Sec:MIL}: Baseline setting).
Second row: localizations by our method, which adds size order to the baseline (purple, see Sec.~\ref{Sec:MIL}: Size order).
Third row: localizations by our method with both size order and weighting (red, see Sec.~\ref{Sec:MIL}: Size weighting).
Ground-truth bounding-boxes are shown in blue.}
\label{Fig:result}
\end{figure}

Table~\ref{Tab:Comparison} shows the result: CorLoc 39.1 on the training set $\iset$  and mAP 20.1 on the test set $\xset$. Examples are in Fig.~\ref{Fig:result} first row. In the following, we incorporate our ideas (size order and size weighting) into this baseline (Alg. 1).

\medskip
\begin{table}[t]
\centering
\caption{\small Comparison between the baseline MIL scheme, various versions of our scheme, and the state-of-the-art on PASCAL 07\label{Tab:Comparison}. `Deep' indicates using additional MIL iterations with Fast R-CNN as detector.
 }
\begin{tabular}{|c||c|c|c||cc|} \hline
	\multicolumn{4}{|c||}{Method}   & CorLoc            &mAP              \\ \hline \hline
		   & size order & size weight     &  deep      &       -       &      -      \\ 
		\hline
				Baseline   &  &   &   &  39.1  & 20.1      \\ 
				\hline
	 \multirow{3}{*}{Our scheme}	  & \cm&                     &       & 46.3              &  24.9          \\
	 	  & \cm &  \cm                    &       & 55.8       &  28.0          \\
	      & \cm &  \cm                & \cm       & \bf{60.9}        &   \bf{36.0}       \\ 
	     \hline 
	     Baseline   &  &   &  \cm  &  43.2  & 24.7      \\ 
	     \hline \hline
	\multicolumn{4}{|c||}{Cinbis~\etal~\cite{cinbis15pami}}&     54.2              & 28.6             \\
	\multicolumn{4}{|c||}{Wang~\etal~\cite{wang15tip}}    &      48.5              & 31.6              \\
		\multicolumn{4}{|c||}{Bilen~\etal~\cite{bilen15cvpr}}    &      43.7              & 27.7              \\
	\multicolumn{4}{|c||}{Shi~\etal~\cite{shi15pami}}     &      38.3              & -              \\
	\multicolumn{4}{|c||}{Song~\etal~\cite{song14nips}}   &     -                  & 24.6              \\ \hline
\end{tabular}
\end{table}

\medskip

\para{Size order.}
We use the same settings as the baseline ($K = 3$ and $M = 3$), but now the training set $\iset$ is split into batches according to the size estimates.
As Table~\ref{Tab:Comparison} shows, by performing curriculum learning based on size order, we improve CorLoc to 46.3 and mAP to 24.9. Examples are in Fig.~\ref{Fig:result} second row.
\medskip

\para{Size weighting.}
Significant improvement of CorLoc can be further achieved by adding size weighting on top of size order. Table~\ref{Tab:Comparison} illustrates this effect: the CorLoc using size order and size weighting goes to 55.8. Compared the the baseline 39.1, this is a $+16.7$ improvement. Furthermore, the mAP improves to 28.0 ($+7.9$ over the baseline). Examples are in Fig.~\ref{Fig:result} third row.
\medskip

\para{Deep net.}
So far, we have used an SVM on top of fixed deep features as the appearance model. Now we change the model to a deeper one, which trains all layers during the re-training step of MIL (Sec.~\ref{Sec:shallow_MIL}). We take the best detection result we obtained so far (using both size order and size weighting) as an initialization for three additional MIL iterations.
During these iterations, we use Fast R-CNN~\cite{girshick15iccv} as appearance model.
We use the entire set at once (no batches) during the re-training and re-localization steps, and omit bounding-box regression in the re-training step~\cite{girshick15iccv}, for simplicity.
We only carry out three iterations as the system quickly converges after the first iteration.

As Table~\ref{Tab:Comparison} shows, using this deeper model raises CorLoc to 60.9 and mAP to 36.0, which is a visible improvement.
It is interesting to apply these deep MIL iterations also on top of the detections produced by the baseline. This yields a $+4.1$ higher CorLoc and $+4.6$ mAP (reaching 43.2 CorLoc and 24.7 mAP). In comparison, the effect of our proposed size order and size weighting is greater ($+16.7$ CorLoc and $+7.9$ mAP over the baseline, when both use SVM appearance models). Moreover, size order and weighting have an even greater  effect when used in conjunction with the deep appearance model ($+17.7$ CorLoc and $+11.3$ mAP, when both the baseline and our method use Fast R-CNN).

\medskip

\para{Comparison to the state-of-the-art.}
Table~\ref{Tab:Comparison} also compares our method to state-of-the-art WSOL works~\cite{cinbis15pami,wang15tip,bilen15cvpr,shi15pami,song14nips}.
We compare both the CorLoc on the trainval set $\iset$ and mAP on the test set $\xset$.
We list the best results reported in each paper.
Note~\cite{cinbis15pami} removes training images with only truncated and difficult object instances, which makes the WSOL problem easier, whereas we train from all images.
%
As the table shows, our method outperforms all these works both in terms of CorLoc and mAP.
All methods we compare to, except~\cite{shi15pami} use AlexNet, pretrained on ILSVRC classification data, as we do. 

\medskip

\subsection{Impact of size of training set for size regressor}
\label{Sec:Details_analysis}

\begin{figure}[t]
 \centering
 \subfigure {\includegraphics[width=0.32\columnwidth]{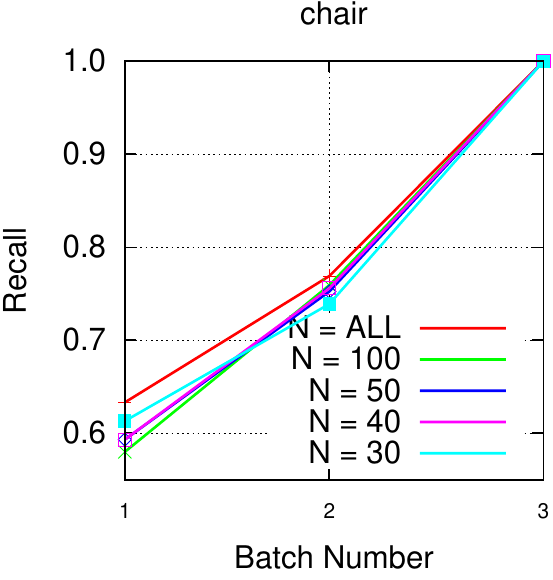}}
 \subfigure{\includegraphics[width=0.32\columnwidth]{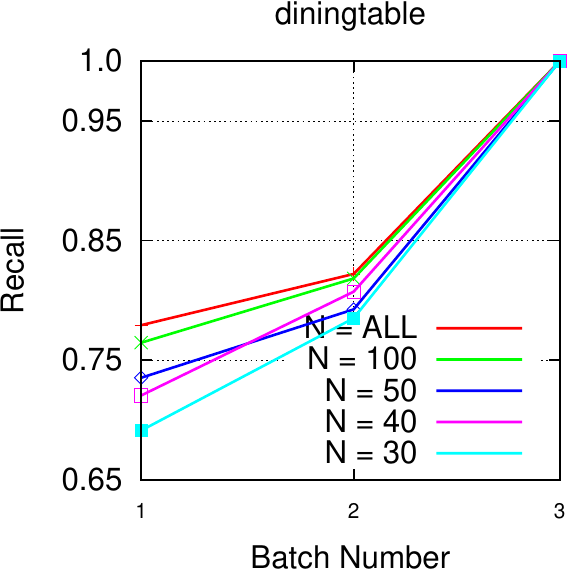}}
  \subfigure{\includegraphics[width=0.32\columnwidth]{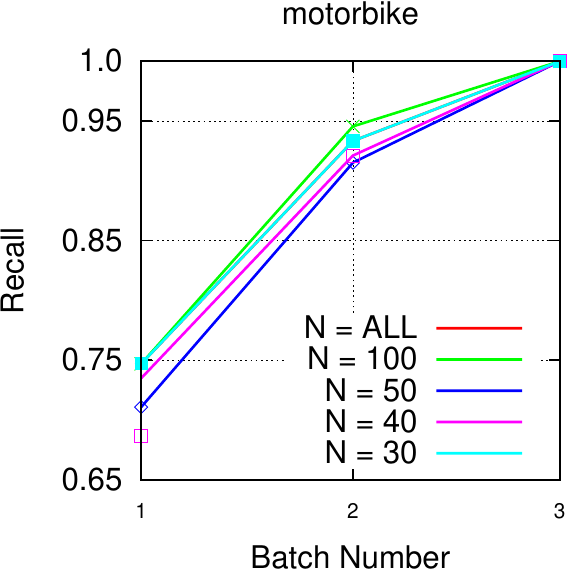}}
 \caption{\small Correlation between inter-batch size order based on the ground-truth size sequence and the estimated sequence, on class \emph{chair}, \emph{diningtable}, and \emph{motorbike} of $\iset$ set; $\mathrm{recall}$ is computed as in (\ref{Eqn: precision-seq}). }
\label{Fig:recall-regression}
\end{figure}

The size estimator we used so far is trained on the complete set $\rset$. What if we only have limited training samples with object size annotations?
As shown in Sec.~\ref{Sec:Preliminary}, when we reduce the number of training samples $N$ per class, the accuracy of size estimation decreases moderately. However, we argue that neither Kendall's $\tau$ nor MSE are suitable for measuring the impact of the size estimates on MIL, when these are used to establish an order as we do in Sec.~\ref{Sec:size_order}.
As $\iset$ is split into batches according to the size estimates, only the inter-batch size order matters, the order of images within one batch does not make any difference.

To measure the correlation of inter-batch size order between the ground-truth size sequence ${Q_{GT}}$ and the estimated size sequence $Q_{ES}$, we count how many samples in ${Q_{GT}^k}$ have been successfully retrieved in ${Q_{ES}^k}$, where $Q^k$ indicates the set of images in batches 1 through $k$:
%

\begin{equation}\label{Eqn: precision-seq}
  \mathrm{recall} = \frac{{|Q_{GT}^k \cap Q_{ES}^k|}}{{|Q_{GT}^k|}},
\end{equation}
$| \cdot |$ denotes number of elements. Fig~\ref{Fig:recall-regression} shows recall curves on set $\iset$, with varying $N$.
The curves are quite close to each other, showing that reducing $N$ does not affect the inter-batch order very much.

\begin{figure}[t]
\center
 \subfigure[CorLoc]{\includegraphics[width=0.49\columnwidth]{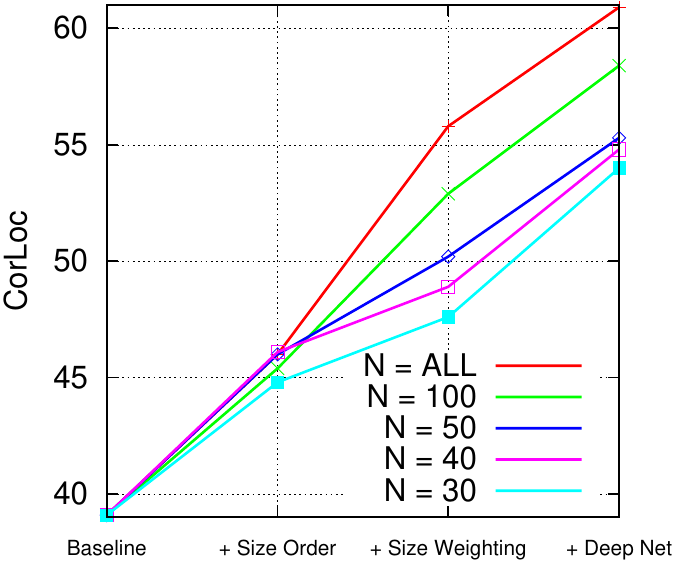}}
 \subfigure[mAP]{\includegraphics[width=0.48\columnwidth]{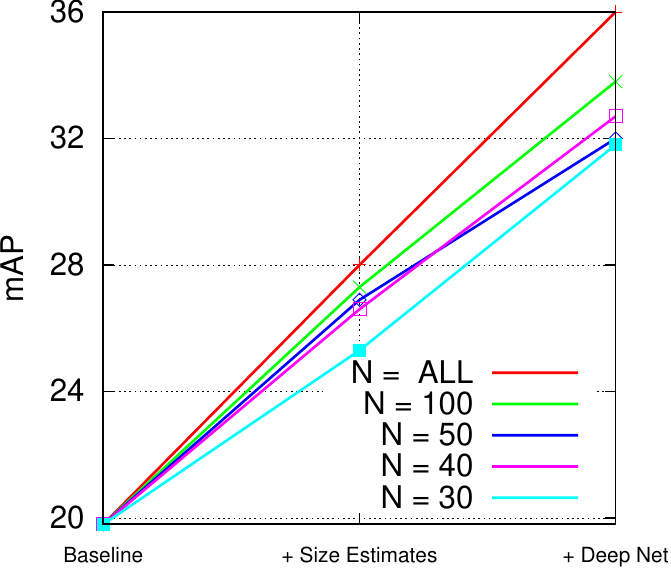}}
\caption{\small WSOL performance on PASCAL 07 when varying $N$. Size order and weighting are gradually added into the baseline MIL framework, and eventually fed into the deep net. We use `size estimates' in (b) to denote using both size order and size weighting.}
\label{Fig:corloc}
\end{figure}

In Fig.~\ref{Fig:corloc} we conduct the WSOL experiment of Sec.~\ref{Sec:MIL}, incorporating size order into the basic MIL framework on $\iset$, using different size estimators trained with varying $N$. The `baseline + size order' result in Fig.~\ref{Fig:corloc}a shows little variation: even $N=30$ leads to CorLoc within 2\% of using the full set $N=\mathrm{ALL}$.
This is due to the fact shown above, that a less accurate size estimator does not affect the inter-batch size order much.

We also propose to use the size estimate to help MIL with size weighting (Sec.~\ref{Sec:size_weighting}).
Table~\ref{Tab:regression} shows that MSE gets larger when $N$ becomes smaller, which means the estimated object size gets father from the real value. This lower accuracy estimate affects size weighting and, in turn, can affect the performance of MIL.
To validate this, we add size weighting on top of size order into MIL in Fig.~\ref{Fig:corloc}.
This time, the CorLoc improvement brought by size weighting varies significantly with $N$.
Nevertheless, even with just $N=30$ training samples per class, we still get an improvement. We believe this is due to the three-sigma rule we adopted in the weighting function (\ref{Eqn:size_score}). The real object size is very likely to fall into the $3\sigma$ range, and so it gets a relatively high weighting compared to the proposals with size outside the range.

Finally, we apply the additional deep MIL iterations presented in Sec.~\ref{Sec:MIL}, `Deep net'. Fig.~\ref{Fig:corloc} shows a consistent trend of improvement across different $N$ and our proposed size order and weighting schemes, on both CorLoc and mAP.

\medskip

\subsection{Further analysis}\label{Sec:further_analysis}

\para{Deep \vs Deeper.}
So far we used AlexNet~\cite{krizhevsky12nips} during deep re-training (Sec:~\ref{Sec:MIL}, `Deep net' paragraph). Here we use an even deeper CNN architecture, VGG16~\cite{simonyan15iclr}. The result in Table~\ref{Tab:CNN} shows the benefits by going deeper, as get to a final CorLoc 64.7 and mAP 37.2.

\medskip

\begin{table}[t]
\caption{\small WSOL results using AlexNet or VGG16 in Fast R-CNN. We report CorLoc on the trainval set $\iset$ and mAP on the test set $\xset$ of PASCAL 07. \label{Tab:CNN}}
\begin{center}
   \begin{tabular}{|c|c|c|}
     \cline{1-3}
   CNN architecture & {AlexNet~\cite{krizhevsky12nips}}&{VGG16~\cite{simonyan15iclr}}\\
     \cline{1-3}
       CorLoc (trainval) & 60.9  &  64.7  \\
   \hline
    mAP (test)&  36.0 & 37.2 \\
   \hline
    \end{tabular}
\end{center}
\end{table}

\para{Class-specific, class-generic and across-class.}
So far we used an object size estimator trained separately for each class. Here we test the class-generalization ability of proposed size order and size weighting ideas.
We perform two experiments.
In the first, we use the entire $\rset$ to train a single size estimator over all 20 classes, and use it on every image in $\iset$, regardless of class. We call this estimator {\em class-generic} as it has to work regardless of the class it is applied to, within the range of classes it has seen during training.
In the second experiment, we separate the 20 classes into two groups:
(i) bicycle, bottle, car, chair, dining table, dog, horse, motorbike, person, TV monitor;
(ii) airplane, bird, boat, bus, cat, cow, potted plant, sheep, sofa, train.
We train two size estimators separately, one on each group. When doing WSOL on a class in $\iset$, we use the estimator trained on the group not containing that class.
We call this estimator {\em across-class}, as it has to generalize to new classes not seen during training.

Table~\ref{Tab:Generic} shows the results of WSOL, in terms of CorLoc on the trainval set $\iset$ and the mAP on the test set $\xset$ of PASCAL 07. Thanks to our robust batch-by-batch design in curriculum learning, the CorLoc using the size order is about the same for all size estimators. This shows that it is always beneficial to incorporate our proposed size order into WSOL, even when applied to new classes.
When incorporating also size weighting into MIL, the benefits gradually diminish when going from the class-specific to the across-class estimators, as they predict object size less accurately. Nonetheless, we still get about $+3$ CorLoc when using the class-generic estimator and about $+1$ when using the across-class one.

The last column of Table~\ref{Tab:Generic}, reports mAP on the test set, with deep re-training. The class-generic estimator leads to mAP 32.2, and the across-class one to 30.0.
They are still substantially better than the baseline (24.7 when using deep re-training, see Table~\ref{Tab:Comparison}).
Interestingly, the across-class result is only moderately worse than the class-generic one, which was trained on all 20 classes. This shows our method generalizes well to new classes.


\vspace{-0.1cm}
\begin{table}[t]
\caption{\small WSOL results using different size estimators. The first four columns show CorLoc on the trainval set $\iset$; the last row shows mAP on the test set $\xset$. The baseline does not use size estimates and is reported for reference. \label{Tab:Generic}}
\begin{center}
   \begin{tabular}{|c|c|c|c|c||c|}
     \cline{1-6}
   Size estimator & Baseline & + Size order & + Size weighting & + Deep net & mAP on test $\xset$\\
     \cline{1-6}
       class-specific & 39.1 & 46.3  &  55.8 &  60.9& 36.0\\
   \hline
       class-generic  & 39.1& 45.6 & 48.4 & 54.4& 32.2\\
   \hline
       across-class   & 39.1& 45.0 & 45.8 & 51.1& 30.0\\
      \hline
    \end{tabular}
\end{center}
\end{table}

\section{Conclusions}

We proposed to use object size estimates to help weakly supervised object localization (WSOL). We introduced a curriculum learning strategy to feed training images into WSOL in an order from images containing bigger objects down to smaller ones. We also proposed to use the size estimates to help the re-localization step of WSOL, by weighting object proposals according to how close their size matches the estimated object size. We demonstrated the effectiveness of both ideas on top of a standard multiple instance learning WSOL scheme.

Currently we use the output of the MIL framework with size order and size weighting as the starting point for additional iterations that re-train the whole deep net. However, the training set is not batched any more during deep re-training. A promising direction for future work is to embed the size estimates into an MIL loop where the whole deep net is updated. Another interesting direction is to go towards a continuous ordering, \ie where the batch size goes towards $1$; efficiently updating the model in that setting is another challenge.

\noindent \textbf{Acknowledgements.} Work supported by the ERC Starting Grant VisCul.

\clearpage

\bibliographystyle{splncs}
\bibliography{bibtex/shortstrings,bibtex/vggroup,bibtex/calvin}

\end{document}